\documentclass[conference]{IEEEtran}
\usepackage{cite}
\usepackage{amsmath,amssymb,amsfonts}
\usepackage{algorithmic}
\usepackage{graphicx}
\usepackage{float}
\usepackage{textcomp}
\usepackage{comment}
\usepackage{xcolor}
\usepackage{pdfpages}
\usepackage{caption}
\usepackage{subcaption}
\usepackage{mathtools}
\usepackage{multirow}
\usepackage{longtable}

\def\BibTeX{{\rm B\kern-.05em{\sc i\kern-.025em b}\kern-.08em
    T\kern-.1667em\lower.7ex\hbox{E}\kern-.125emX}}
    
\makeatletter
\newcommand{\linebreakand}{%
  \end{@IEEEauthorhalign}
  \hfill\mbox{}\par
  \mbox{}\hfill\begin{@IEEEauthorhalign}
}
\makeatother

\begin{document}    

\title{Exploring Unsupervised Learning Methods for Automated Protocol Analysis\\

}


\author{\IEEEauthorblockN{Arijit Dasgupta}
\IEEEauthorblockA{\textit{Mechanical Engineering} \\
\textit{National University of Singapore}\\
Singapore, Singapore \\
arijit.dasgupta@u.nus.edu}

\and
\IEEEauthorblockN{Yi-Xue Yan}
\IEEEauthorblockA{\textit{Electrical Engineering} \\
\textit{Nanyang Technological University}\\
Singapore, Singapore \\
yany0025@e.ntu.edu.sg}

\and
\IEEEauthorblockN{Clarence Ong}
\IEEEauthorblockA{\textit{Data Science and Analytics} \\
\textit{National University of Singapore}\\
Singapore, Singapore \\
clarence\_ong@u.nus.edu}

\linebreakand
\IEEEauthorblockN{Jenn-Yue Teo, Bugsy}
\IEEEauthorblockA{\textit{Electronic Systems Division} \\
\textit{DSO National Laboratories}\\
Singapore, Singapore \\
tjennyue@dso.org.sg}

\and
\IEEEauthorblockN{Dr Chia-Wei Lim, Andrew}
\IEEEauthorblockA{\textit{Electronic Systems Division} \\
\textit{DSO National Laboratories}\\
Singapore, Singapore \\
lchiawei@dso.org.sg}
}


\maketitle

\begin{abstract}
The ability to analyse and differentiate network protocol traffic is crucial for network resource management to provide differentiated services by Telcos. Automated Protocol Analysis (APA) is crucial to significantly improve efficiency and reduce reliance on human experts. There are numerous automated state-of-the-art unsupervised methods for clustering unknown protocols in APA. However, many such methods have not been sufficiently explored using diverse test datasets. Thus failing to demonstrate their robustness to generalise.

This study proposed a comprehensive framework to evaluate various combinations of feature extraction and clustering methods in APA. It also proposed a novel approach to automate selection of dataset dependent model parameters for feature extraction, resulting in improved performance. Promising results of a novel field-based tokenisation approach also led to our proposal of a novel automated hybrid approach for feature extraction and clustering of unknown protocols in APA. 

Our proposed hybrid approach performed the best in 7 out of 9 of the diverse test datasets, thus displaying the robustness to generalise across diverse unknown protocols. It also outperformed the unsupervised clustering technique in state-of-the-art open-source APA tool, NETZOB in all test datasets. 


\end{abstract}


\section{Introduction}

Communication protocols are a predefined set of rules that multiple parties use at different OSI layers to communicate consistently. Despite there being many open-standards protocols (e.g. TCP, IP \& 802.11), there are still numerous proprietary unknown protocols owned by companies \& organizations. Therefore, there is the need for Protocol Analysis (PA) to infer detailed specifications of unknown protocols for network resource management, IoT interoperability, network protocol security audit, simulation and conformance testing \cite{duchene2018state}. Furthermore, the ability to analyse and differentiate network protocol traffic at routers (especially those of unknown protocols) is vital for effective network resource management by Telcos for differentiated Quality of Service (QoS).

This paper focuses on PA via Static Traffic Analysis based on analysis of captured network traffic of unknown protocols. This is a two stage process, where: 1) Vocabulary inference involves understanding the protocol messages, and 2) Grammar inference involves understanding the protocol predefined set of rules. Vocabulary inference involves clustering protocol messages into smaller and similar groups for subsequent inference of protocol field boundaries, relationships and semantics.

Traditionally, PA is done manually by experts and is very time-consuming, taking months or even years, with the additional challenges of having to recruit, train and retain such experts. Therefore, the need for APA was first raised in \cite{cui2007discoverer}, with proposed approaches inspired by disciplines such as Bioinformatics and Natural Language Processing (NLP), due to the likeness in sequential semantics derived from byte sequences in protocol packets \cite{krueger2010asap, antunes2011reverse, netzob}. This has significantly improved the efficiency of the analysis process and even reduce the reliance on human experts.  Today, NETZOB \cite{netzob} is the most comprehensive open-source APA framework that utilizes a bioinformatics-inspired method for protocol message clustering \cite{beddoe2004network} and is our baseline for comparison.


This paper focuses on automated feature extraction and clustering of packets with similar message formats in the vocabulary inference stage. Once this is done, the human expert analyst can analyse packets in each cluster more easily and efficiently. Hence, our framework aids the human expert analyst in PA and provides a crucial and early step affecting subsequent stages of the APA pipeline. We assume no prior knowledge and explore various unsupervised methods. The key contributions are as follows:
\begin{enumerate}
    \item Developed a comprehensive APA framework for evaluation of various combinations of state-of-the-art unsupervised feature extraction \& clustering methods.
    \item Proposed novel methods for automated model optimisation for APA and developed greater insights into techniques for automatic field-based tokenization.
    \item Comprehensive experimentation of unsupervised automated features extraction and unknown protocol message clustering for APA, leading to an improved hybrid approach over state-of-the-art open-source APA tool, NETZOB and other related works.
\end{enumerate}

Section~\ref{relatedworks} presents related works, Section~\ref{framework} presents our proposed APA framework and methods, Section~\ref{experiments} describes the experiment methodology, Section~\ref{results} discusses our experiment results and finally Section~\ref{conclusion} concludes the paper.

\section{Related Works}
\label{relatedworks}
Related works have typically focused on inferring message format types from packets of a single unknown protocol \cite{netzob} - \cite{luo2019type}, using feature extraction and clustering techniques such as sequence alignment \cite{netzob} and information bottleneck \cite{wang2012semantics}. Today, there are hundreds of different protocols and it is naive and limiting to assume that a stream of unknown packets belong to a single protocol. Hence, our proposed framework (in Section~\ref{framework}) aims to automatically differentiate both unknown protocol and message format types via distinguishing features to facilitate further analysis in later APA stages.

Previous works have also typically used information from the entire packet for feature extraction \cite{netzob} - \cite{luo2019type}. However, only the header of protocol packets usually contain information with relevance to the protocol's operation. Hence, it is desirable to perform feature extraction on only the packet header. Faced with an unknown protocol, there is no information on the length of this header portion. Therefore, our framework introduces a novel method that aims to infer the header length of the unknown protocol that yields the most amount of useful information for differentiating unknown protocols. 

To extract features of an unknown protocol, sequence alignment techniques from bioinformatics and NLP have been employed due to the similarities in structure of DNA sequences and packets in network traces, and the textual nature of packet contents. Bossert et al. proposed NETZOB \cite{netzob}, which uses Needleman-Wunsch Sequence Alignment (NWSA) from bioinformatics to infer message formats and cluster protocols \& message types, while Discoverer \cite{cui2007discoverer} uses tokenisation, recursive clustering and merging clusters to do so. Both techniques require expert knowledge on common delimiters in the protocol. To account for unknown protocols, the framework in the present study does not rely on common delimiters or any form of expert knowledge unlike these existing methods. The global sequence alignment technique NWSA used by NETZOB is also computationally time expensive ($\mathcal{O}(n^2)$), where $n$ is the number of data packets), and makes use of only observable literal information, ignoring semantic information. Luo et al \cite{luo2019type} proposed using Latent Dirichlet Allocation (LDA) from NLP to study the type distributions derived from the statistics of message N-grams to infer protocol message formats of different types. However, the study did not perform an extensive hyper-parameter tuning of the $\alpha$ \& $\beta$ that control the Dirichlet distribution or to select the size of LDA topics.

Kleber et al proposed NEMESYS \cite{kleber2018nemesys} that uses the delta of the congruence in bit values of consecutive bytes to identify field boundaries in a packet using its intrinsic message structure. For text-based protocols with longer fields, this intrinsic structure-aware approach of obtaining field boundaries is able to produce more meaningful tokens than n-grams. NEMESYS is used by in novel field-based tokenisation approach.

\section{Proposed APA Framework}
\label{framework}
The proposed APA framework comprises steps shown in Fig \ref{fig:EnigmaFramework}. It does not strictly mandate a linear application of methods, but rather encapsulates a set of unsupervised methods to be potentially used in combination. Section~\ref{exptsetup} lists the combination of methods that we evaluated. Methods in each step of the framework are described as follows:

\begin{figure}[h!]
\centering
\includegraphics[trim={0 3cm 1 3cm},clip,width=\columnwidth]{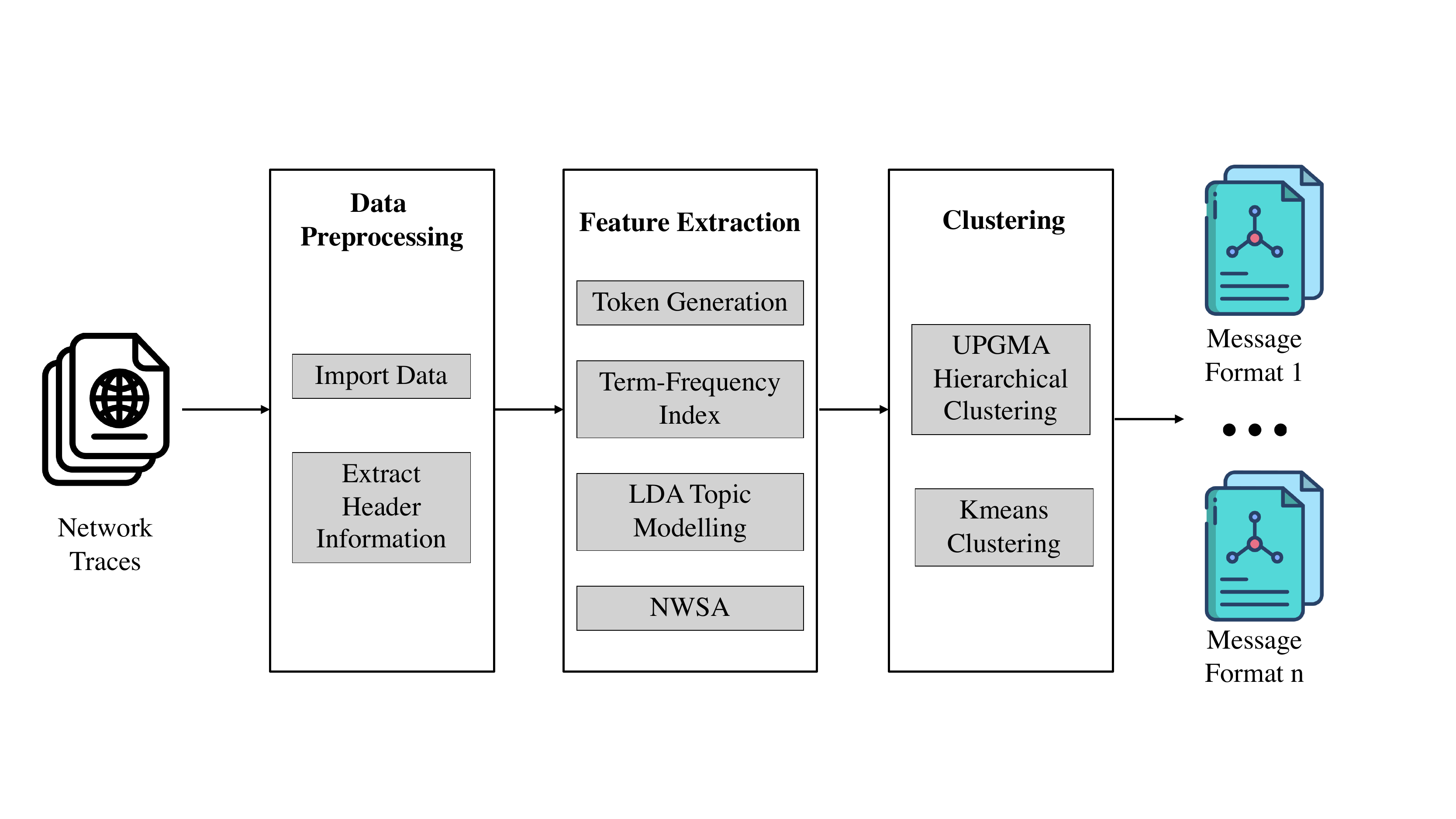}
\caption{Overview of Proposed APA Framework}
\label{fig:EnigmaFramework}
\end{figure}

\subsection{Data Pre-Processing}
\label{preprocess}
Network traces in PCAP format are imported using the Python SCAPY library as binary or hexadecimal packets to facilitate feature extraction. We assumed that all protocols at OSI layers below the unknown protocol are known and their headers are therefore stripped from the packet. Finally, the header of the unknown protocol can be extracted from the remaining packet using the inferred header length from Section~\ref{inferhdrlen}. Since many application layer protocols do not have a header, we do not extract a header from the application layer and instead, use the remaining payload.

\subsection{Feature Extraction}
\label{fe}
The framework proceeds to extract distinguishing features, from the header of the unknown protocol, that will be utilized in subsequent clustering stage to differentiate packets from different protocols and message format types.

\subsubsection{Tokens Generation}
In NLP's N-grams tokenisation \cite{wang2012semantics}, one gram is a single word and adjacent words in a text string are combined to form tokens. In contrast, protocol headers are typically parsed as binary strings, one gram is represented by a single byte and consecutive bytes combined to form a N-grams token. Alternatively for text-based protocols with longer fields, a novel field-based tokenisation approach is proposed where NEMESYS \cite{kleber2018nemesys} is used to infer field boundaries within a protocol header and tokenisation performed along these boundaries. The resultant corpus of tokens generated for each protocol header packet can be further analysed by advanced statistical methods like LDA in Section~\ref{lda} to generate distinguishing features for subsequent clustering.

\subsubsection{Term-Frequency (TF) Index}
With a generated corpus of tokens per protocol packet, the next step is to generate an appropriate feature representation that is distinct for different protocols and message formats. First, we explored using the TF index for feature extraction. By recording the raw count of unique tokens in each corpus, a TF matrix of size $p\times n$ is generated (where $p$ is number of protocols and $n$ is number of unique tokens). Despite its intuitive representation, the generated matrix is often sparse, making computational storage unnecessarily expensive. Furthermore, the high dimensionality of the matrix meant that the application of dimensional reduction methods such as Principal Component Analysis (PCA) would often be required as clustering performance generally depreciates with higher dimensions.

\subsubsection{Latent Dirichlet Allocation (LDA)}
\label{lda}
Due to the limitations of the TF index discussed above, we sought a more efficient alternative. Next, we explored the use of LDA from NLP for feature extraction. LDA is an unsupervised topic modelling approach that allocates the generated tokens to a set of predefined number of LDA topics, based on the statistical extent of dissimilarity in which each individual token shared with other tokens in the corpus. Using the  structured topic modelling (stm) package in R \cite{roberts2019stm}, a vector representation for each protocol packet in the data set is generated. This vector represents the posterior probability of the packet belonging to a particular LDA topic, given the corpus of tokens.

\subsubsection{Optimising LDA Topic Size}
\label{lda_optim}
The topic size is a key LDA hyper-parameter that determines not only the size, but also the generated feature representation that will be used to represent each protocol packet of a dataset. As such, it is crucial for practical deployment to automate and optimise the selection of this key LDA hyper-parameter, as optimising the topic size results in better clustering performance, with the optimised value being specific to the individual dataset. However despite its importance, such hyper-parameter tuning has not been explored in previous works. 

By utilising the mean semantic coherence and exclusivity scores of a LDA topic size \cite{roberts2019stm} as unsupervised metrics, the LDA topic size hyper-parameter can be automatically optimised for a given dataset, thus resulting in better clustering performance. 
The FREX metric \cite{bischof2013nd} is used as a measure to quantify the degree of exclusivity of a given topic in a way that balances the word frequency, with FREX\textsubscript{k,v} being the weighted harmonic mean of the rank of token v in topic k (Equation \ref{eq:equation1}).
While the exclusivity score provides a quantity of measure for the degree of dissimilarity between LDA topics generated from a specific size, the semantic coherence score measures the extent in which tokens in the same topic co-occur together in the same communication protocol (Equation \ref{eq:equation2}).

\begingroup
\footnotesize
\begin{equation}
\label{eq:equation1}
FREX\textsubscript{k,v} = \Bigg(\frac{\omega}{ECDF(\beta_{k,v} / \sum_{j=1}^{K}{\beta_{j,v}})} + \frac{1-\omega}{ECDF(\beta_{k,v})} \Bigg) ^{-1}
\end{equation}
\normalsize
where ECDF is the empirical CDF, ${\beta_{k,v}}$ is the topic-specific frequency of token v in topic k and $\omega$ is the weight set to 0.7 to favor exclusivity.
\endgroup

\begingroup
\footnotesize
\begin{equation}
\label{eq:equation2}
C\textsubscript{k} = \sum_{i=2}^{M}{\sum_{j=1}^{i-1}{log \Bigg( \frac{D(v\textsubscript{i},v\textsubscript{j}) + 1)}{D(v\textsubscript{j})}} \Bigg)}
\end{equation}
\normalsize
where C\textsubscript{k} is the semantic coherence for topic k, D(v\textsubscript{i},v\textsubscript{j}) is the number of times tokens v\textsubscript{i} and v\textsubscript{j} appear together in the same protocol and D(v\textsubscript{j}) is the total number of times the token v\textsubscript{j} appears in the data set.
\endgroup \\\

\begin{figure}[h!]
  \includegraphics[scale= 0.5, width=\columnwidth]{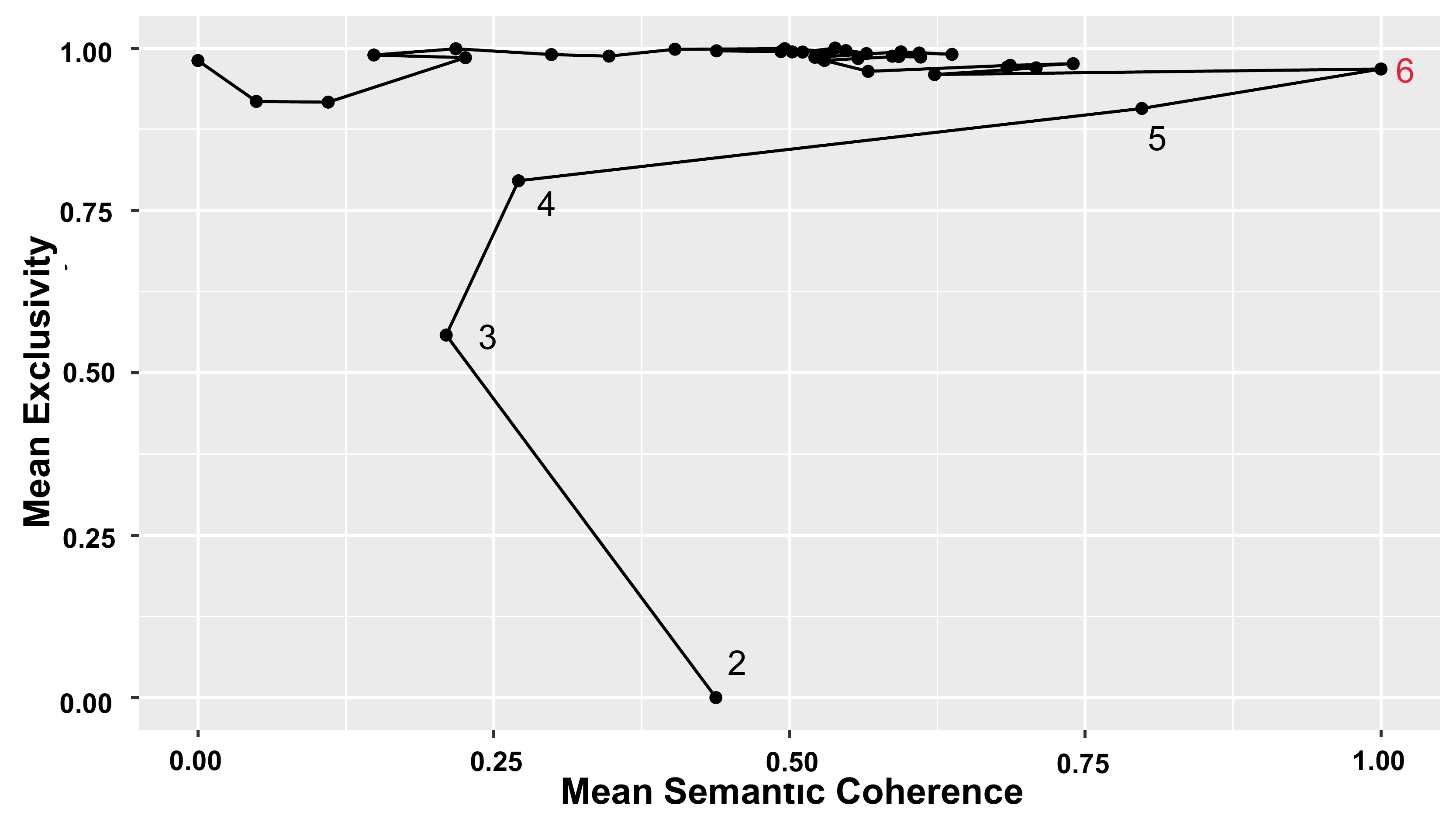}
  \caption{Mean Exclusivity against Mean Semantic Coherence for Link Layer dataset. Points represent mean values for a topic size, with topic size increasing along the trace.}
  \label{fig:Topic_Size_Link}
\end{figure}

With the derivation of the exclusivity and semantic coherence scores for each LDA topic in a given size of generated topics, the mean values of the topics in each size were used as a measure for the overall quality of the topics generated from that specific topic size. With the vast difference in scales of the mean exclusivity and semantic coherence scores, the values were normalised to between 0 and 1. The optimum topic size can then be determined graphically by the data point with the greatest Euclidean distance from the origin. This data point will correspond to the selected optimised LDA topic size hyper-parameter. For example, in Figure \ref{fig:Topic_Size_Link}, it is observed that the optimum number of LDA topics selected for the Link Layer dataset (described in Section~\ref{dataset}) was 6.

\subsubsection{Optimising Extracted Protocol Header Length}
\label{inferhdrlen}
In the data pre-processing stage (Section~\ref{preprocess}), we would require the length of the unknown protocol header to be accurately estimated, as this would result in the right amount of features to be extracted, in order to obtain good clustering results. Similar to optimising the LDA topic size hyper-parameter (Section~\ref{lda_optim}), optimising the extracted header length in the pre-processing is also crucial for practical deployment that has not yet been explored in previous works.

Based on our analysis, key features found in the header of the protocol packets are often sufficient in distinguishing between the different protocol and message format types.
Conversely, using the entire protocol packet for feature extraction and clustering would often introduce a significant amount of stochastic noise into the data, thus adversely affecting the quality of LDA topics generated and subsequently clustering performance.
Therefore, it is crucial to be able to accurately estimate the appropriate protocol header length to be extracted, and used for subsequent feature extraction and clustering stages. Just like the LDA topic size, it was observed that the extracted header length was also a hyper-parameter value that was specific to the individual data set.

We extend the approach to optimise the LDA topic size hyper-parameter (Section~\ref{lda_optim}) by varying the header length, and produced different iterations of the plot in Figure \ref{fig:Topic_Size_Link}. 
The goal is to determine the appropriate header length to be used for a dataset, given the various plots generated. A novel approach proposed, is to select the header length that generated the plot with the most isolated optimal data point (i.e. highest euclidean distance from origin). 
Consequently, the optimised header length selected would generate the optimised LDA topic size with greatest difference in exclusivity and semantic coherence scores. Mathematically, the degree of isolation is measured by mean difference in Euclidean distance between the optimum data point and its two adjacent neighbours.

With the combined h yper-parameter search space of the LDA topic size and the protocol header length having a modestly small area, we were able to iterate through all permutations of the 2 hyper-parameters, in order to optimize for the best APA performance 

\subsubsection{Needleman-Wunsch Sequence Alignment (NWSA)}
NSWA from bioinformatics is used by NETZOB \cite{needleman1970general} to compute the alignment score, by comparing the similarity between packet sequences via global sequence alignment. Iterating over each dataset, a matrix of alignment scores can be generated for clustering. It is  more expensive computationally and thus unsuitable for clustering of large datasets.

\subsection{Unsupervised Clustering}
The framework proceeds to cluster packets of the same protocol or message format into disjoint sets using the set of representative features extracted via methods in Section~\ref{fe} in an unsupervised manner via a similarity metric.

\subsubsection{Similarity Metric}
Conventionally, determining the degree of similarity between data points has often been through the use of Euclidean distance in the N-dimensional subspace. However, given that the dimension of the set of representative features increases with LDA topic size, the degree of sparseness also increases exponentially. Therefore, the curse of dimensionality \cite{koppen2000curse} makes Euclidean distance a poor similarity metric candidate for protocols clustering. 
Alternatively, the Cosine similarity metric is often a better choice for high-dimensional data that considers each data point to be represented by a single vector and scores the similarity between pairwise vectors based on the angle between them.


\subsubsection{UPGMA Hierarchical Clustering}
\label{upgma}
\begin{figure}[h!]
  \centering
  \includegraphics[scale= 0.34, trim={0 5cm 0cm 0cm},clip, width=\columnwidth]{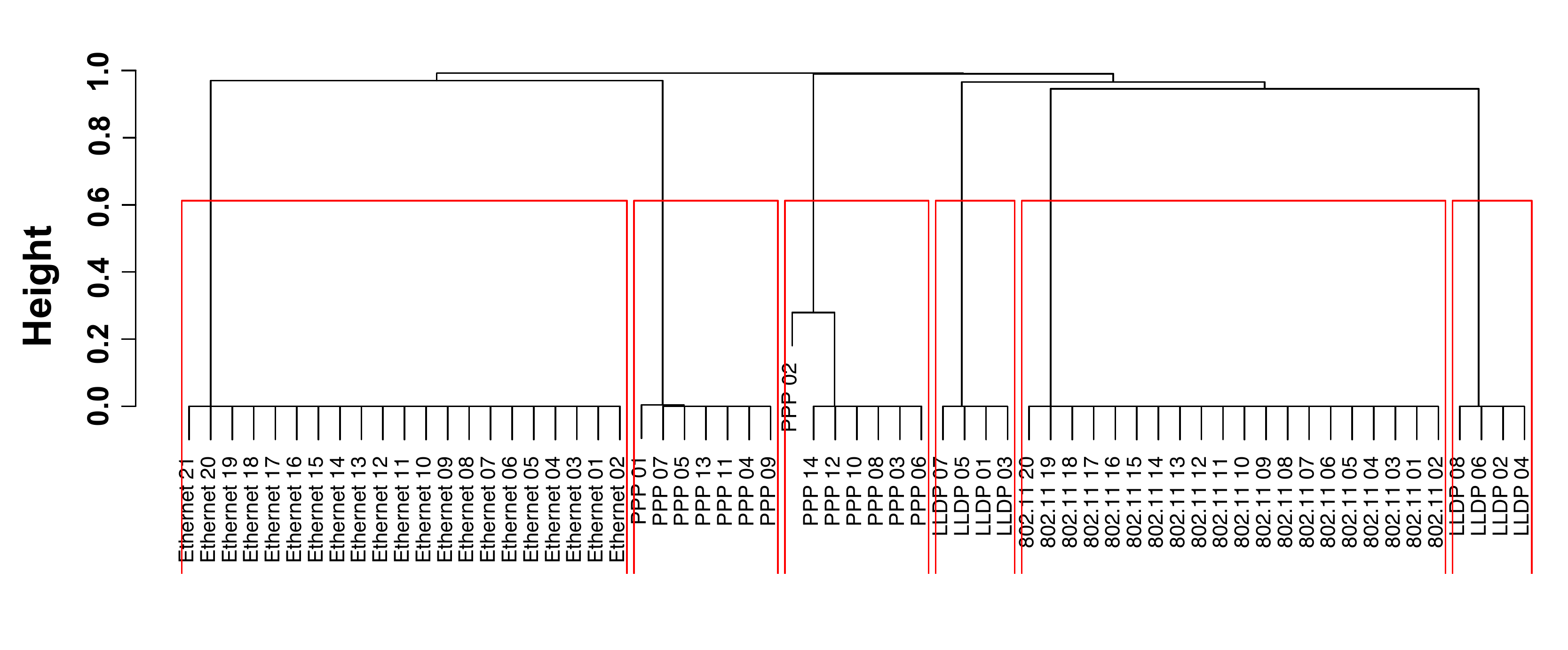}
  \caption{Example of a dendrogram with 6 disjoint clusters for Link Layer Protocols.}
  \label{fig:Dendrogram}
\end{figure}

The unweighted pair group method with arithmetic mean (UPGMA) algorithm is an agglomerative hierarchical clustering algorithm that regards each protocol packet as first belonging to a single cluster and then proceeds to combine the 2 most similar clusters to form a larger overarching cluster in an iterative manner using the cosine similarity metric. This is repeated until the threshold to indicate the minimum degree of dissimilarity between the clusters is exceeded. We use a static threshold of 0.5 and this results in a dendrogram exemplified in Figure \ref{fig:Dendrogram}. 

\subsubsection{K-Means Clustering}
Alternatively, the K-Means clustering algorithm is an unsupervised learning algorithm that first assigns the data points randomly into $K$ distinct and disjoint clusters, and then iteratively shifts the $K$ centroids based on the class association of the data points to the nearest centroid. In each step, the algorithm assigns the data points to new clusters, such that the within sum of squares of the clusters are reduced in each iteration, eventually converging to a global minimum at some point. The number of predefined centroids, K, can be determined through the "elbow method", which is automated with the use of the Kneedle algorithm ~\cite{kneedle}. The stochastic nature of the clustering output is due to the random initialisation of the $K$ centroids that may not be desirable.

\section{Experiments}
\label{experiments}
\subsection{Datasets}
\label{dataset}
This study aims to group unknown protocols with similar packet formats at different OSI layers, and group packets of an unknown protocol with similar message formats. Hence, a wide variety of protocol and message format types is necessary to test the framework extensively. The diversity with and within our 9 datasets in total, comprehensively evaluates the robustness and the ability of our proposed framework to generalise for unknown protocols and achieve the aforementioned aims. Each dataset comprises 200 protocol packets. These 200 packets are selected by performing stratified sampling on a larger dataset (see Appendix) for each of the 9 datasets. The packets from the dataset detailed in the appendix are sourced from open-source databases, mainly WireShark Wiki. It is also assumed that the traffic is not encrypted.

We specifically engineered two characteristics of the dataset to make clustering more challenging as these characteristics are meant to replicate the exacting realities of un-encrypted network traces. First, we intentionally used a maximum of only 200 packets per dataset. This is because clustering with less data simulates scenarios with limited data, and so performing better with less data would further highlight the robustness of our framework. Second, the dataset is unbalanced like most un-encrypted network traffic. This additional step makes it challenging to cluster rare message types. In a supervised problem, designing a dataset with the two mentioned characteristics would generally be seen as a weakness as data-driven models thrive on more ground-truth based data. As the present study presents an unsupervised learning problem, these characteristics instead test the robustness of our framework and shows its ability to perform well under challenging \& real-world circumstances.

5 of the 9 datasets are used for fine-grain type clustering within protocols. The 5 protocols used here are ICMP, TCP, SCTP, DNS and HTTP and the aim is to cluster similar message format types in each protocol. Additionally, we have 4 OSI layers-based datasets, with each dataset containing protocols from the respective OSI layers. The aim is to group protocols with similar packet formats in each layer (e.g. TCP ACK, TCP SYN etc. for the TCP dataset). 
We also chose to split application layer protocols into textual and binary protocols due to the obvious difference in data structure used in these packets. Hence, the two types of protocols may require different sets of hyper-parameters in our proposed APA framework. The technique for differentiating the two types of application layer protocols is based on domain knowledge and involves searching for special ASCII patterns that occurs in textual protocols but not binary protocols. This method of searching using a predefined rule is very lightweight and and has negligible computational cost. 
Note that all message format types in the single protocol-based datasets and all protocols in the OSI layers-based datasets are assumed to be unknown.  Table~\ref{table:datasetTable} describes the protocols, assumed to be unknown, that comprise the 4 OSI layers-based datasets, and some of the message types, assumed to be unknown, that comprise the 5 fine-grain type clustering datasets. 

\begin{table}[]
\caption{Dataset protocol/type description}
\label{table:datasetTable}
\begin{tabular}{ll}
\hline
Dataset                                                                           & Protocols/Types                                                                                                                                                                                           \\ \hline 
\begin{tabular}[c]{@{}l@{}}Link Layer \\ Protocols\end{tabular}                   & \begin{tabular}[c]{@{}l@{}}Point to Point Protocol (PPP), \\ Link Layer Discovery Protocol (LLDP), \\ IEEE 802.11, Ethernet\end{tabular}                                                                  \\ \hline 
\begin{tabular}[c]{@{}l@{}}Transport\\ Layer Protocols\end{tabular}               & \begin{tabular}[c]{@{}l@{}}Internet Control Message Protocol (ICMP), \\ Transmission Control Protocol (TCP), \\ User Datagram Protocol (UDP), \\ Stream Control Transmission Protocol (SCTP)\end{tabular} \\ \hline 
\begin{tabular}[c]{@{}l@{}}Application \\ Layer Protocols\\ (Text)\end{tabular}   & \begin{tabular}[c]{@{}l@{}}Domain Name Server (DNS), \\ Routing Information Protocol (RIP), \\ Transport Layer Security (TLS)\end{tabular}                                                                \\ \hline 
\begin{tabular}[c]{@{}l@{}}Application \\ Layer Protocols\\ (Binary)\end{tabular} & \begin{tabular}[c]{@{}l@{}}Trivial File Transfer Protocol (TFTP),\\ Hypertext Transfer Protocol (HTTP),\\ Simple Mail Transfer Protocol (SMTP)\end{tabular}                                               \\ \hline 
\begin{tabular}[c]{@{}l@{}}TCP Message \\ Types\end{tabular}                      & \begin{tabular}[c]{@{}l@{}}7 TCP message types \\ e.g. ACK, PSH ACK, SYN, RST, FIN ACK\end{tabular}                                                                                                       \\ \hline 
\begin{tabular}[c]{@{}l@{}}SCTP Chunk \\ Types\end{tabular}                       & \begin{tabular}[c]{@{}l@{}}16 SCTP chunk types\\ e.g. INIT, COOKIE ECHO, DATA, \\ HEARTBEAT, ASCONF, ACK\end{tabular}                                                                                     \\ \hline 
ICMP Types                                                                        & \begin{tabular}[c]{@{}l@{}}4 ICMP types\\ e.g. Reply, Request, Destination Unreachable, \\ TTL Exceeded\end{tabular}                                                                                      \\ \hline 
HTTP Methods                                                                      & \begin{tabular}[c]{@{}l@{}}3 HTTP methods\\ e.g. 200 OK, GET, POST\end{tabular}                                                                                                                           \\ \hline 
\begin{tabular}[c]{@{}l@{}}DNS message \\ types\end{tabular}                      & \begin{tabular}[c]{@{}l@{}}4 DNS message types\\ e.g. Query, Response Refused, Response \\ No Error, Response No Such Name\end{tabular}                                                                   \\ \hline 
\end{tabular}
\end{table}

\subsection{Experimental Setup}
\label{exptsetup}
Our experiments have three objectives. First, to compare the different tokenisation methods of proposed framework, which is done by comparing N-grams and NEMESYS~\cite{kleber2018nemesys} across the 9 datasets. Second, to evaluate both the proposed LDA topic size and extracted protocol header length optimisation methods. This is done by varying the LDA topic sizes and header lengths to observe how the chosen topic size and header length compare with the actual best-performing topic size and header length. And finally, to compare the overall clustering performance across 5 different combinations of feature extraction \& clustering methods of the proposed framework. 

The first  is to use the open source APA tool, NETZOB that utilizes NWSA feature extraction and UPGMA clustering. The second combines LDA features extraction with K-MEANS clustering. The third combines LDA features extraction with UPGMA clustering.The fourth uses TF index feature extraction with UPGMA clustering. N-grams tokenisation is used with all feature extraction methods, with the exception of NWSA.  Finally, the fifth is our proposed hybrid approach that automatically selects the best feature extraction method based on our findings from the previous four approaches. By default, TF index feature extraction is used with UPGMA clustering. However, for application layer binary protocols, LDA feature extraction is used, and for application layer textual protocols, our proposed field-based tokenisation based on NEMESYS is used instead. The detection of binary or textual protocols can be achieved automatically via suitable predefined rules.

We repeat the entire process of proposed APA framework (in Section~\ref{framework}) for all 9 datasets. Before starting our experiments, a comprehensive hyper-parameter tuning process was done. We first conducted a sensitivity analysis on which hyper-parameters affected the performance more and sorted them in a hierarchical list. Afterwards we sampled the more important hyper-parameters via grid search and went down the list. Note that as the header length and LDA topic size were dataset-sensitive, they were not fixed via hyper-parameter tuning, but they were determined using the proposed method in Sections~\ref{lda_optim} \& \ref{inferhdrlen}. For NEMESYS, the $\sigma$ value for bit congruence was optimal at 0.5 and that no tokens should be kept longer than 40 bytes. For N-grams tokenisation, hexadecimal packet representation was more efficient than binary for all feature extraction methods. Moreover, a gram size of 3 bytes was determined to be optimal. Optimal values for all tuned hyper-parameters will be used across all datasets. 


All computations were run on a Windows 10 machine with Intel(R) Core(TM) i7-8550U CPU processor @1.80GHz and 16GB RAM. Computation effort of each run of our proposed APA framework for each dataset is strongly dependent on the dataset and ranges between 42.25$s$ and 357.62$s$ with a mean of 138.18$s$. Run-time difference between the techniques compared were negligible for each dataset.

\subsection{Performance Metrics}

To quantify the overall clustering performance, the nature of the output must be realised. After clustering, the data packets are grouped together in clusters, but each cluster is not augmented with a class label due to the nature of this problem being unsupervised. For instance, if 100 data packets (of which 50 are TCP and 50 are UDP) are run through the APA framework and the output is 2 clusters (one of size 45 and another 55), the clusters can only be labelled '1' and '2' (cluster labels) but not TCP or UDP (class labels) directly. This lack of association is why traditional classification metrics like Accuracy and F-score cannot be used as metrics.

However, one may propose an additional step to convert cluster labels into class labels via a voting algorithm. A majority voting algorithm can be employed for each cluster by labelling it with the class label determined by the most common ground truth label in each cluster. For instance, a cluster with 40 packets, with 10 of them being UDP and 30 being TCP, would wholly be labelled as a TCP cluster with the UDP packets considered mis-clustered with the TCP packets. With this, a confusion matrix can be generated and the standard classification metrics can be determined. One flaw of this voting method is that it is swayed by any dataset bias. If the cluster of 45 packets comes from a dataset of 10\% UDP and 90\% TCP, then the cluster would have proportionately more UDP packets and hence become classified as UDP wholly. The voting method can therefore be improved to a proportion-based majority voting algorithm. After further review, the present study abstained from using any voting algorithm for two reasons.

First, the voting algorithm is an artificial step and does not do justice to the nature of clustering. Clustering only groups data points together, it does not label data points or groups. It would be more sensible to use metrics that measured how well similar data points are grouped together and how dissimilar data points are grouped separately without the need for class labels for every cluster. Second, the hyper parameters can be tampered with to artificially boost the accuracy score by taking advantage of another flaw of the voting algorithm. By setting the number of clusters to be very high in K-Means (or an extremely low threshold for UPGMA), each data packet can be forced to be in its own cluster, forcing the class label to always be correct. This would output an accuracy of 1, even though the output from the APA framework is nonsensical and of no use to an analyst. On the other end of the spectrum, the APA framework can be made to put all data packets into one cluster. This would mean that a dataset with 70\% TCP packets would be labelled with an accuracy of 0.7 even though the output is also nonsensical to an analyst.

To circumvent the issue of having no class labels, the present study looked at few extrinsic clustering metrics; namely the Adjusted Rand Index (ARI) \cite{ari_metric}, the Fowlkes Mallows Score (FMS) \cite{fowlkes1983method} \& the Adjusted Mutual Information (AMI) \cite{ami_score}. All three metrics compare between two clusters, which would be the output (in terms of cluster labels) and the ground truth (in terms of class labels). They compare how similarly the two clusters are grouped and adjust for chance. There is a lack of research to support which metric is better given the nature of clusters, hence the present study chose ARI as the main performance metric given its effectiveness \& wide use in literature for unsupervised learning \cite{javed2018community, bioinformatics, yang2016temporal}.

\begin{figure}[H]
     \centering
     \includegraphics[width=0.8\columnwidth]{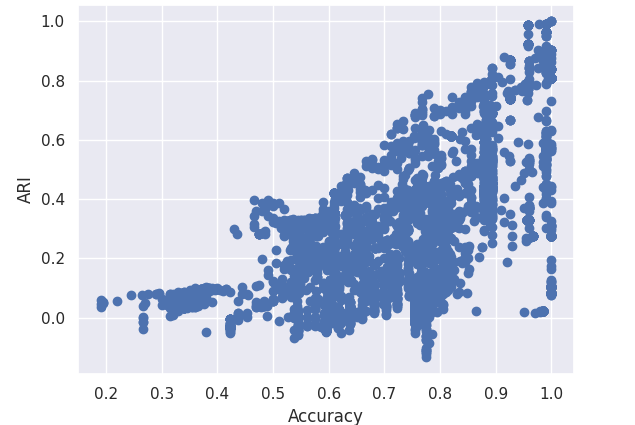}
     \caption{Comparison of ARI against Accuracy with voting in over 13,000 test instances of the APA framework}
     \label{ari_acc}
\end{figure}

Finally, a comparison between ARI \& accuracy with voting is made to illustrate the weakness of the latter metric and determine a threshold for a satisfactory performance using ARI. During all of the experimental testing instances of the APA framework ($>$13,000), the ARI and accuracy with voting results were gathered and compared as shown in Figure~\ref{ari_acc}. The thousands of data points on the bottom right quadrant shows all instances where the clustering performance was poor (ARI), yet the accuracy with voting was indicated to be high. The limitations of accuracy with voting forces the value to be higher. Hence, a high value may not indicate good clustering, but a low value generally indicates bad clustering. Based on Figure~\ref{ari_acc}, any ARI over 0.4 never produces an accuracy lower than 0.6. It is difficult to determine a proper threshold of ARI for a satisfactory outcome for it is not as intuitive as accuracy. Hence, we use an ARI value of 0.4 as the threshold as it has been shown that any value of ARI below 0.4 could be associated with bad clustering (low accuracy).

\section{Results and Discussion}
\label{results}

\label{resanaly}
\begin{figure}[h]
     \centering
     \includegraphics[width=\columnwidth, trim=0 2cm 0 0]{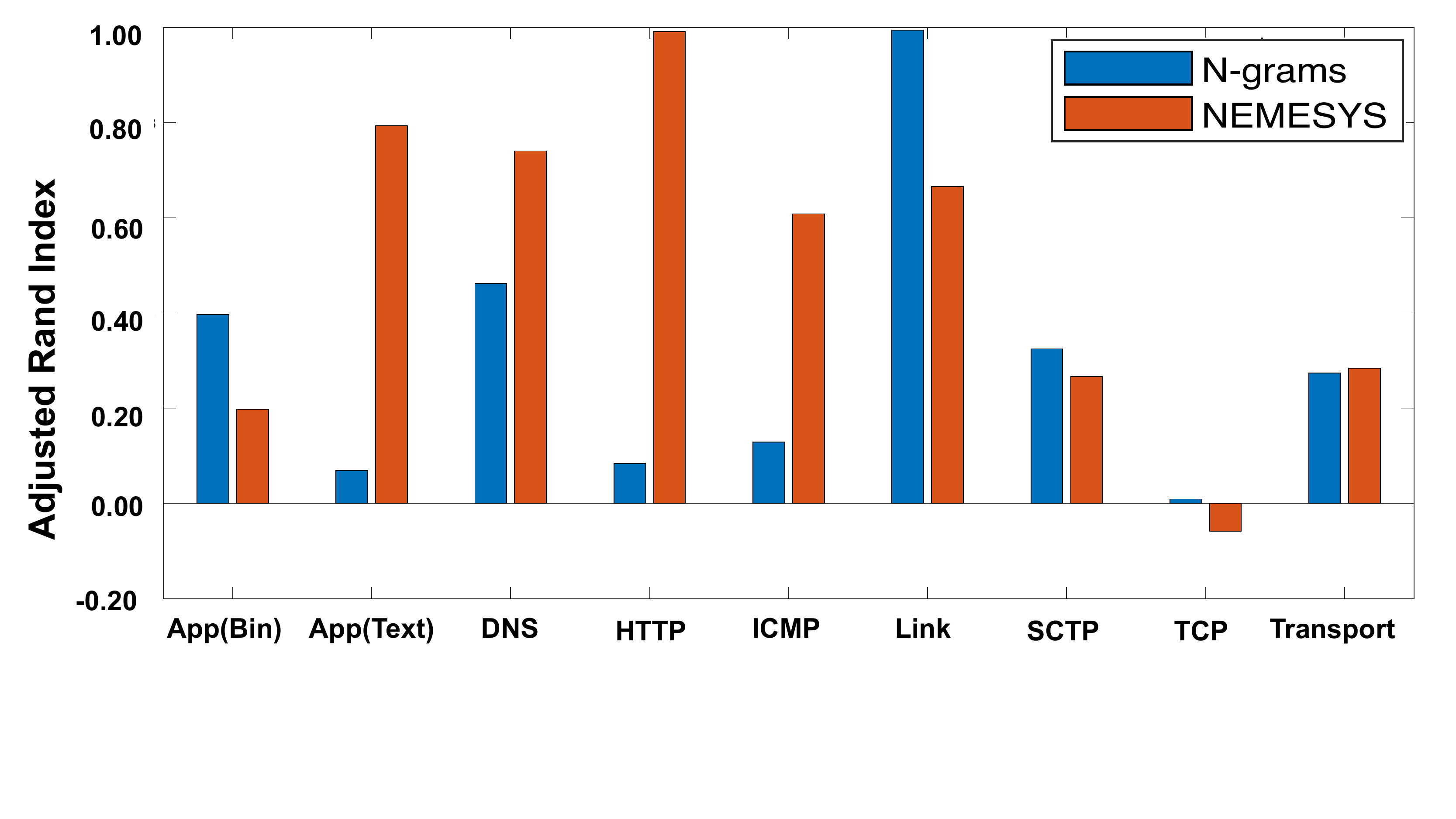}
     \caption{ARI of NEMESYS vs. N-grams tokenisation}
     \label{tokenisation}
\end{figure}

Our first objective is to compare the tokenisation methods (Section~\ref{exptsetup}). Figure~\ref{tokenisation} shows the ARI for NEMESYS and N-grams tokenisation across the 9 datasets. There is no clear winner and NEMESYS out-performs in 5 out of 9 datasets. Closer analysis shows that NEMESYS performs significantly better than N-grams tokenisation for both application layer textual protocols and HTTP protocol datasets, which is also an application layer textual protocol. Our results suggest that NEMESYS tokenization is more effective for application layer textual protocols. This leads us to propose a novel field-based tokenisation based on NEMESYS for application layer textual protocols in our proposed hybrid approach (Section~\ref{exptsetup}).

\begin{figure}[h]
     \centering
     \includegraphics[trim={0 0cm 0cm 0cm},clip, width=0.8\columnwidth]{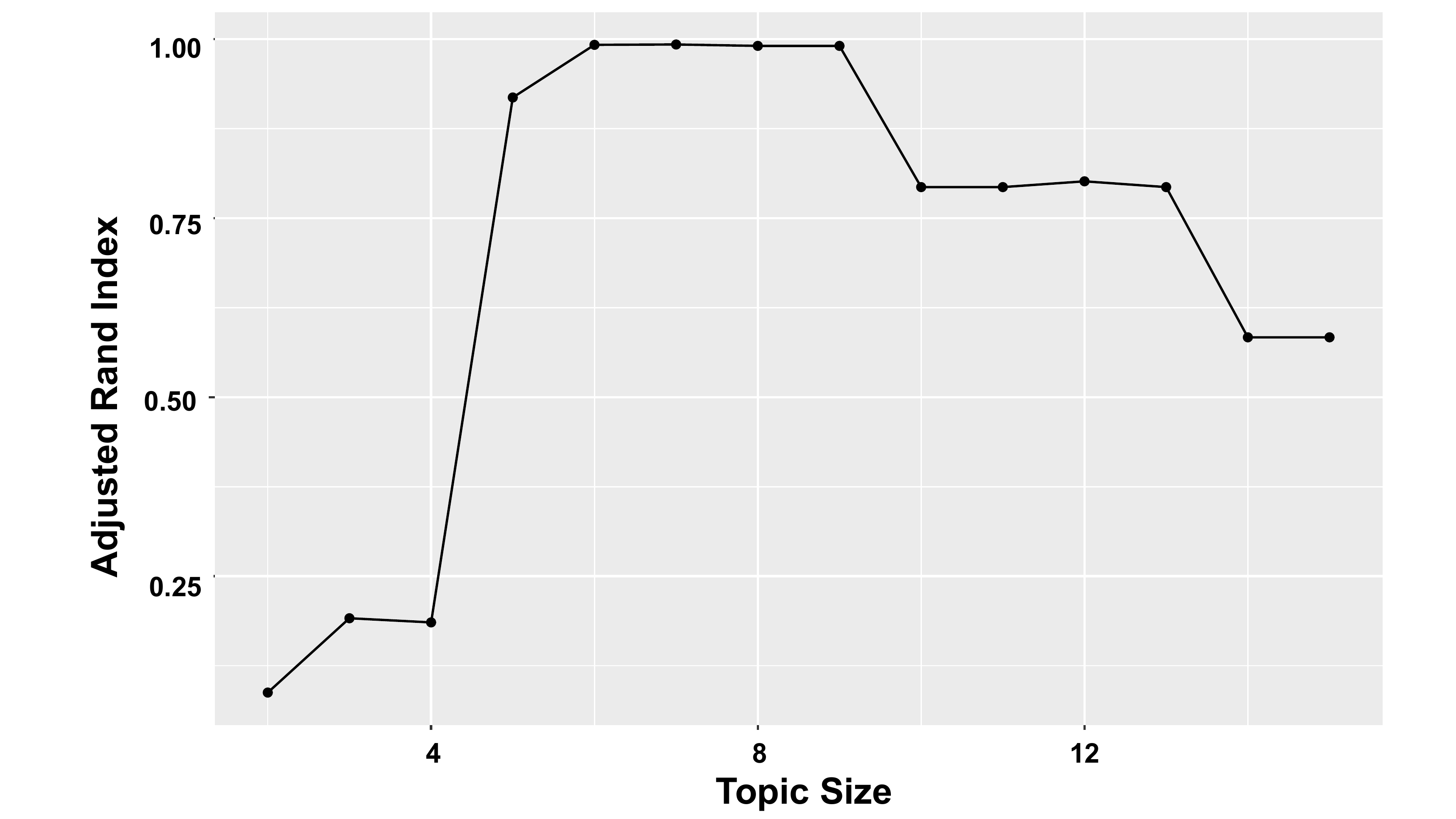}
     \caption{ARI of LDA feature extraction against topic size}
     \label{linktopicsize}
\end{figure}

Our second objective is to evaluate both the proposed LDA topic size and extracted protocol header length optimisations. 
Figure~\ref{fig:Topic_Size_Link} shows that the optimised LDA topic size for the link layer dataset was chosen to be 6.
For validation, we plotted how ARI varies with changing topic size (Figure~\ref{linktopicsize}) and observed that performance was poor (i.e. low ARI) with small topic sizes, but improves until ARI peaks with optimal topic size 6. Similarly, the LDA topic sizes selected using proposed optimisation method in Section~\ref{lda_optim} are either optimal or near-optimal for the other datasets.

\begin{figure}[h!]
     \centering
     \begin{subfigure}[h]{0.49\linewidth}
         \centering
         \includegraphics[trim=13cm 0 0 0, clip=true, width=\linewidth]{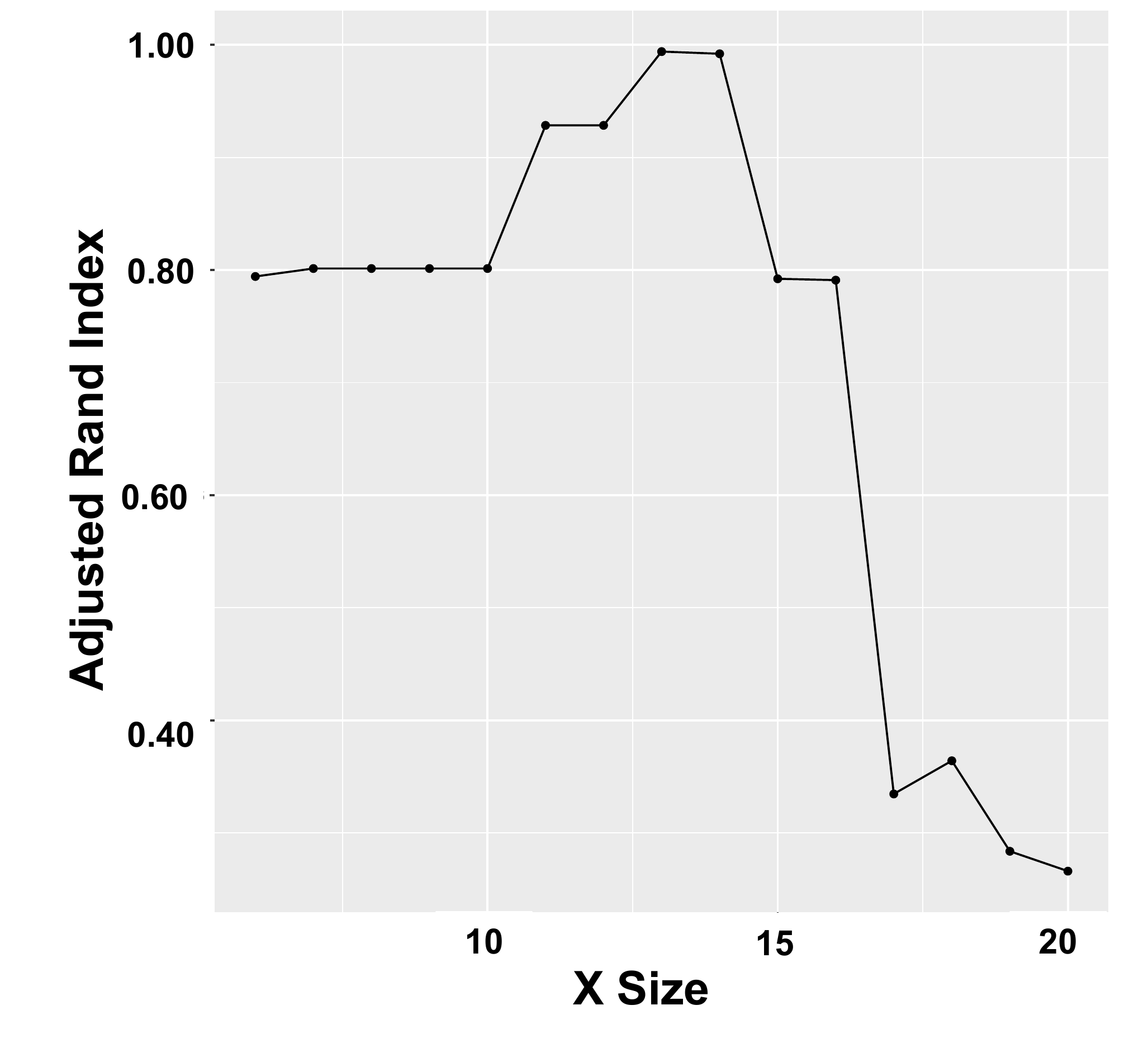}
         \caption{Link Layer dataset}
         \label{link_headerlen}
     \end{subfigure}
     \hfill
     \begin{subfigure}[h]{0.49\linewidth}
         \centering
         \includegraphics[trim=13cm 0 0 0, clip=true,  width=\linewidth]{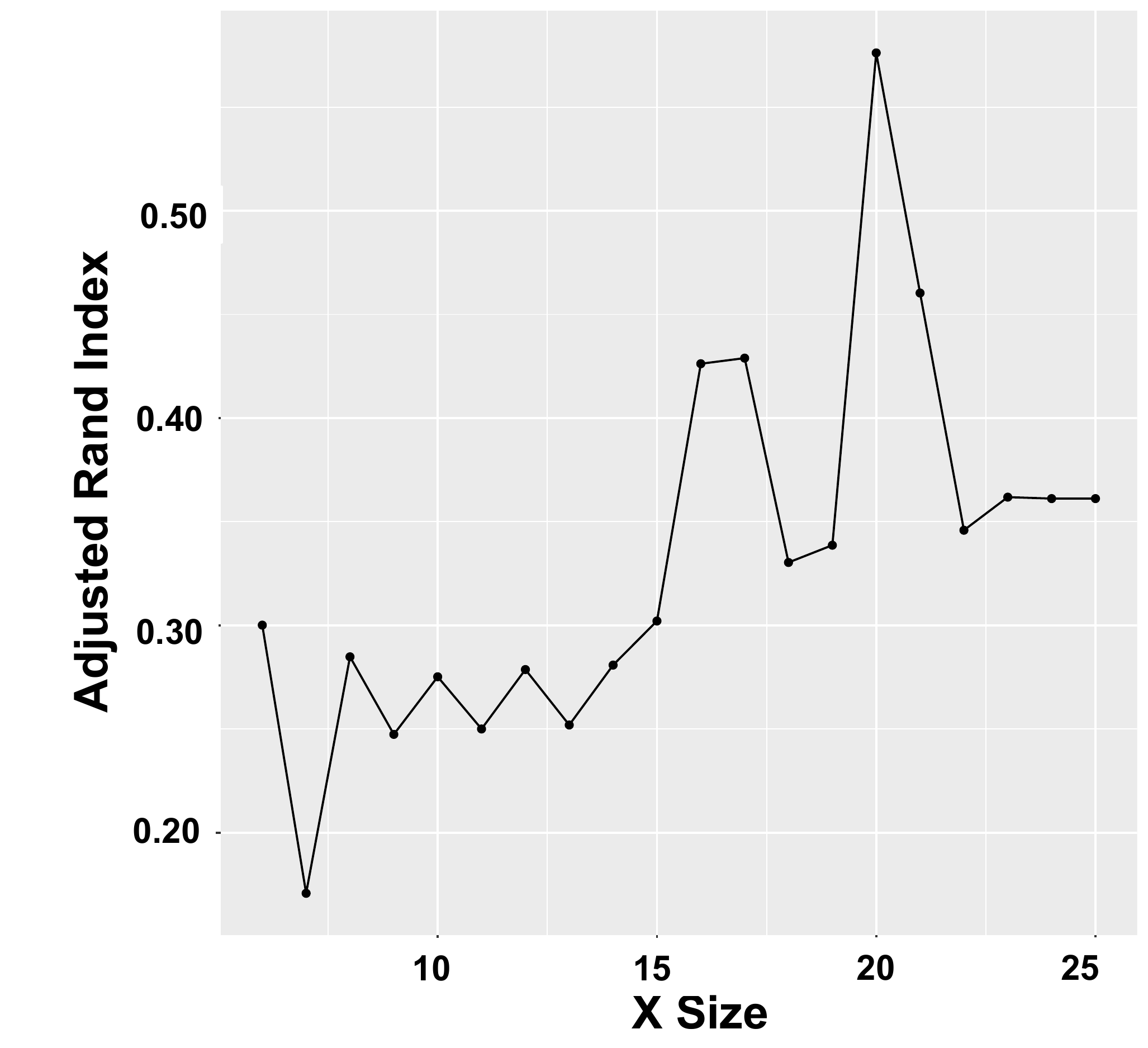}
         \caption{Transport Layer dataset}
         \label{tpt_headerlen}
     \end{subfigure}
        \caption{ARI of LDA feature extraction against protocol header length for Link and Transport datasets}
        \label{headerlen}
\end{figure}

For optimising the extracted protocol header length needed in pre-processing stage (Section~\ref{preprocess}), Figures \ref{link_headerlen} \& \ref{tpt_headerlen} show that the optimised protocol header lengths for link layer and transport layer datasets were chosen to be at 14 and 20 bytes respectively, which corresponded to highest ARI scores. Similarly, the header lengths selected using proposed optimisation method in Section~\ref{inferhdrlen} were either optimal or near-optimal for the other datasets.
Therefore, both proposed LDA topic size and extracted protocol header length optimisations have been validated and are deemed crucial for practical deployment, which has not yet been previously explored.

\begin{figure}[h]
     \centering
     \includegraphics[width=\columnwidth]{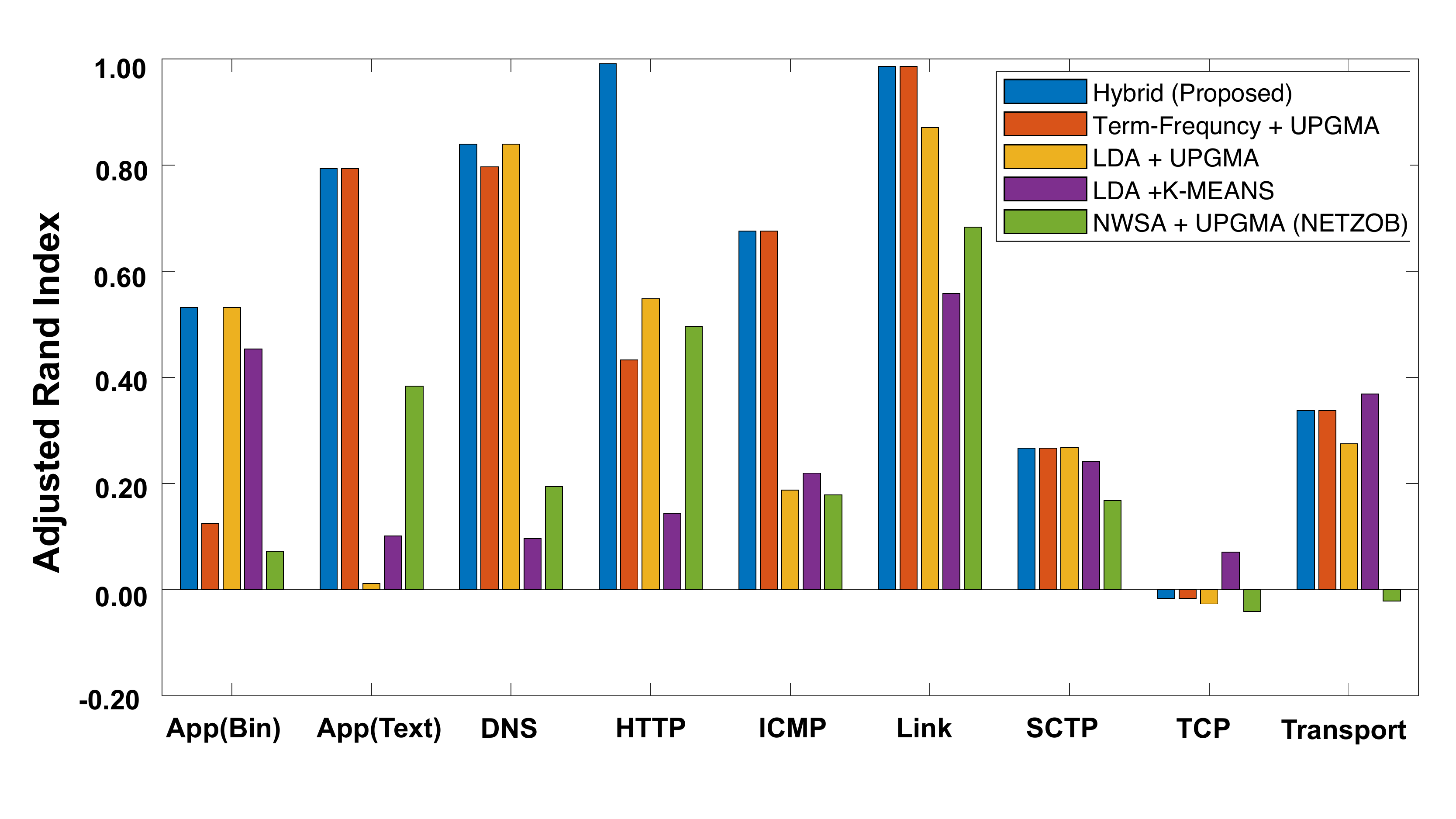}
     \caption{Comparing ARIs of 5 approaches across 9 datasets}
     \label{overall}
\end{figure}

Figure~\ref{overall} compares final clustering performance of the 5 approaches (Section~\ref{exptsetup}) across all 9 datasets. Results show that our proposed hybrid approach is best performing in 7 out of 9 datasets, with ARI $>$ 0.4 for 6 datasets. Unfortunately, all approaches did not achieve satisfactory performance (ARI $<$ 0.4) for STCP, TCP and Transport layer datasets. Upon further investigation, we realised that these 3 datasets comprised of more protocols and message format types to differentiate with less distinct features, thus making clustering challenging. However, after further analysis of the UPGMA dendrograms generated from our proposed approach, we observed that by optimising the static UPGMA threshold of 0.5 (Section~\ref{upgma}), it is possible to significantly improve the performance, which we leave for future works. 

Finally, due to the extensive coverage of our 9 datasets across all the OSI layers with diverse protocols and message format types, we have proven the robustness of our proposed hybrid approach to generalise for unknown protocols, which is crucial for practical deployment and has not been adequately addressed in previous works.

\section{Conclusion and Future Works}
\label{conclusion}
In conclusion, we proposed a comprehensive APA framework and evaluated various combinations of feature extraction and clustering methods, including those used by NETZOB \cite{netzob}. 
Our proposed hybrid approach, that utilizes a novel field-based tokenisation based on NEMESYS for application layer textual protocols, is best performing in 7 out of 9 datasets with ARI $>$ 0.4 for 6 datasets. This result proves the robustness and generalising ability of our proposed hybrid approach.
We also validated our proposed automated optimisation methods, for both  LDA topic size and extracted protocol header length, that is crucial for practical deployment.
However, since computational cost was not the primary focus of our present study, more work can be done to optimise our code and this will help us make more detailed comparisons and analysis of the computational costs of our proposed framework and other existing frameworks in future. 

We also hope that our works contributed as crucial foundation stones for future APA works to be built upon. With recent advances in Deep Learning, like Deep Auto-Encoders for automated features extraction, it will be exciting to explore the application of these advanced Machine Learning (ML) methods for unsupervised learning in APA. Finally, we have only explored the tip of the APA iceberg and in the future, we hope to build upon our proposed APA framework to explore application of advanced ML methods in more areas of APA.


\section*{Acknowledgment}
We wish to thank Mr Chia Yong Kang and Mr Tan Ping Liang for their invaluable contributions and feedback. 


\bibliographystyle{IEEEtran}
\bibliography{references}

\section*{Appendix}

\begin{table}[H]
\centering
\caption*{Dataset Support Description}
\label{tab:dataset_values}
\begin{tabular}{lll}
\hline
Dataset &
  Protocols/Types &
  Support \\ \hline
\multirow{4}{*}{\begin{tabular}[c]{@{}l@{}}Link \\ Layer \\ Protocols\end{tabular}} &
  Point to Point Protocol &
  14 \\
 &
  Link Layer Discovery Protocol &
  8 \\
 &
  IEEE 802.11 &
  86 \\
 &
  Ethernet &
  78 \\ \hline
\multirow{4}{*}{\begin{tabular}[c]{@{}l@{}}Transport\\ Layer \\ Protocols\end{tabular}} &
  \begin{tabular}[c]{@{}l@{}}Internet Control Message \\ Protocol\end{tabular} &
  22 \\
 &
  Transmission Control Protocol &
  100 \\
 &
  User Datagram Protocol &
  26 \\
 &
  \begin{tabular}[c]{@{}l@{}}Stream Control Transmission \\ Protocol\end{tabular} &
  38 \\ \hline
\multirow{3}{*}{\begin{tabular}[c]{@{}l@{}}Application \\ Layer Protocols \\ (Binary)\end{tabular}} &
  Domain Name Server &
  38 \\
 &
  Routing Information Protocol &
  12 \\
 &
  Transport Layer Security &
  20 \\ \hline
\multirow{3}{*}{\begin{tabular}[c]{@{}l@{}}Application \\ Layer Protocols\\ (Text)\end{tabular}} &
  Trivial File Transfer Protocol &
  20 \\
 &
  Hypertext Transfer Protocol &
  19 \\
 &
  Simple Mail Transfer Protocol &
  28 \\ \hline
\multirow{7}{*}{\begin{tabular}[c]{@{}l@{}}TCP \\ Message \\ Types\end{tabular}} &
  ACK &
  3357 \\
 &
  PSH ACK &
  348 \\
 &
  SYN &
  315 \\
 &
  SYN ACK &
  288 \\
 &
  RST &
  2 \\
 &
  RST ACK &
  3 \\
 &
  SIN ACK &
  157 \\ \hline
\multirow{15}{*}{\begin{tabular}[c]{@{}l@{}}SCTP \\ Chunk \\ Types\end{tabular}} &
  INIT &
  2 \\
 &
  INIT ACK &
  2 \\
 &
  COOKIE ECHO &
  2 \\
 &
  COOKIE ACK &
  2 \\
 &
  DATA &
  120 \\
 &
  SACK DATA &
  1 \\
 &
  SACK &
  108 \\
 &
  SHUTDOWN &
  3 \\
 &
  SHUTDOWN ACK &
  2 \\
 &
  SHUTDOWN COMPLETE &
  2 \\
 &
  HEARTBEAT &
  73 \\
 &
  HEARTBEAT ACK &
  63 \\
 &
  HEARTBEAT ACK DATA &
  1 \\
 &
  ASCONF &
  3 \\
 &
  ASCONF ACK &
  3 \\ \hline
\multirow{4}{*}{\begin{tabular}[c]{@{}l@{}}ICMP \\ Types\end{tabular}} &
  Reply 0 &
  23 \\
 &
  Request 8 &
  27 \\
 &
  TTL Exceeded 11 &
  12 \\
 &
  Destination Unreachable 3 &
  2 \\ \hline
\multirow{3}{*}{\begin{tabular}[c]{@{}l@{}}HTTP \\ Methods\end{tabular}} &
  200 OK &
  146 \\
 &
  GET &
  537 \\
 &
  POST &
  6 \\ \hline
\multirow{4}{*}{\begin{tabular}[c]{@{}l@{}}DNS \\ message \\ types\end{tabular}} &
  Query &
  36 \\
 &
  Response Refused &
  1 \\
 &
  Response No Error &
  23 \\
 &
  Response No Such Name &
  6 \\ \hline
\end{tabular}
\end{table}

\end{document}